\documentclass[10pt,twocolumn,letterpaper]{article}

\usepackage{cvpr}      
\usepackage{graphicx}
\usepackage{colortbl}
\usepackage{xcolor}
\usepackage{booktabs}
\definecolor{cvprblue}{rgb}{0.21,0.49,0.74}
\usepackage[pagebackref,breaklinks,colorlinks,allcolors=cvprblue]{hyperref}
\usepackage{xcolor}
\def\paperID{10724} 
\def\confName{CVPR}
\def\confYear{2025}

\title{Vision-Language Embodiment for Monocular Depth Estimation}

\author{Jinchang Zhang\\
Intelligent Vision and Sensing Lab\\
University of Georgia\\
{\tt\small jz23267@uga.edu}
\and
Guoyu Lu\\
Intelligent Vision and Sensing Lab\\
University of Georgia\\
{\tt\small guoyulu62@gmail.com}
}

\begin{document}
\maketitle
\begin{abstract}
Depth estimation is a core problem in robotic perception and vision tasks, but 3D reconstruction from a single image presents inherent uncertainties. Current depth estimation models primarily rely on inter-image relationships for supervised training, often overlooking the intrinsic information provided by the camera itself.
We propose a method that embodies  the camera model and its physical characteristics into a deep learning model, computing embodied scene depth through real-time interactions with road environments. The model can calculate embodied scene depth in real-time based on immediate environmental changes using only the intrinsic properties of the camera, without any additional equipment. By combining embodied scene depth with RGB image features, the model gains a comprehensive perspective on both geometric and visual details. Additionally, we incorporate text descriptions containing environmental content and depth information as priors for scene understanding, enriching the model's perception of objects. This integration of image and language — two inherently ambiguous modalities — leverages their complementary strengths for monocular depth estimation. The real-time nature of the embodied language and depth prior model ensures that the model can continuously adjust its perception and behavior in dynamic environments. Experimental results show that the embodied depth estimation method  enhances model performance across different scenes.
\end{abstract}    
\section{Introduction}
Monocular depth estimation is a critical task in robotic perception and computer vision, widely used in autonomous driving, augmented reality \cite{tang2022perception}, and 3D reconstruction \cite{newcombe2011kinectfusion}. Its objective is to assign a depth value to each pixel in an image. However,the mapping from three-dimensional (3D) to two-dimensional (2D) space is inherently ill-posed, depth estimation faces intrinsic ambiguity, requiring identification of the most plausible scene among multiple possibilities.
In this context, we propose a depth estimation method based on language and camera model embodying, integrating the physical characteristics of the camera model into the deep learning system. We compute Embodied Road Depth in real-time according to immediate changes in the road environment and extend it to the entire scene, referred to as Embodied Scene Depth. This method combines the intrinsic properties of the camera and the actual scene, providing reliable depth priors without additional hardware. By using Embodied Scene Depth fused with RGB image features as input for depth estimation, the model can capture geometric priors and visual details of the scene, achieving more accurate depth estimation.
\begin{figure*}[h]
\begin{center}
\includegraphics[width=17cm, height=7cm]{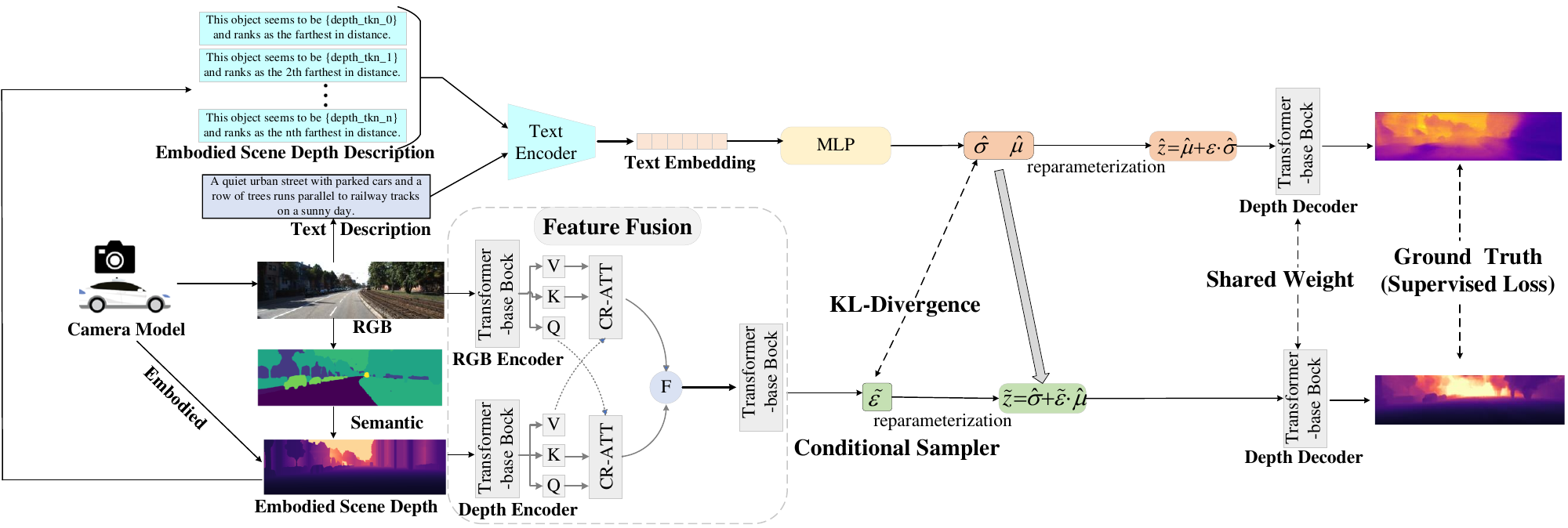}
\end{center}
\vspace{-7mm}
\caption{Overview of the framework. We utilize a plug-and-play pre-trained image segmentation model to obtain segmentation results from images and incorporate the camera model to calculate embodied scene depth. We extract the textual semantic description of the image and derive object depth descriptions based on semantic segmentation and embodied scene depth, merging them into a textual description. The text encoder is used to predict the mean and standard deviation of the latent distribution corresponding to the depth map of the textual description. We then sample $\hat{z}$ from the distribution using the reparameterization trick, where $\epsilon \sim N(0,1)$, and decode it into a depth map for loss computation. In the feature fusion module, we extract features from the embodied scene depth and RGB image and use a cross-attention mechanism for feature fusion. Next, we optimize a conditional sampler by predicting patch-wise $\tilde{\epsilon}$ from the fused features to sample $\tilde{z}$ from the latent space, and output the depth through the depth decoder. The text and image depth decoders share weights and are updated in both alternating steps.
}
\vspace{-6mm}
\label{architecture}
\end{figure*}

Additionally, we embody environmental text descriptions as language priors for scene understanding. The object information embodied in the text descriptions helps constrain the typical sizes and layouts of objects in the scene. Although these descriptions may correspond to multiple compatible 3D scenes, they provide priors that enable the model to capture object scale more effectively. In monocular depth estimation, we leverage the complementary nature of images and text: images provide direct observations of the 3D scene, while text introduces strong priors on object scale and structure. By using an image caption generator (e.g., \cite{hu2022expansionnet}), we generate scene descriptions, and based on semantic segmentation and Embodied Scene Depth, we can obtain depth descriptions and relative positions for each object. These descriptions are integrated as textual priors.
We use a text variational autoencoder (VAE) composed of a CLIP text encoder and an MLP to encode the text into a latent distribution of scene layouts, represented by mean and standard deviation. We use reparameterization to sample from the latent distribution and generate depth maps through a depth decoder. To obtain depth maps corresponding to specific images, we use a conditional sampler to sample from the image, estimating the latent distribution from the text encoding. The resulting latent vector of the scene layout is used by the depth decoder to generate the most probable depth map within the distribution.

In this paper, we detail our contributions, which revolve around the embodying of the camera model and vision-language integration. These contributions have significant, multifaceted impacts on the field of depth estimation based on vision-language models:
1: We propose a method that embodies the camera model and its physical characteristics into a deep learning model to compute embodied scene depth through real-time interactions with road environments, providing reliable priors for depth estimation.
2: We extract depth text descriptions based on embodied scene depth and semantic segmentation, integrating image semantic descriptions, and enable the text variational encoder to learn the distribution of possible corresponding 3D scenes as priors.
3: We introduce a conditional sampler based on multimodal features that fuses embodied scene depth and RGB image features. The conditional sampler samples from the fused features and uses text descriptions as conditional priors for modeling.
4: We propose an embodied depth estimation framework that embodies the camera, environment, and language, leveraging the complementary strengths of image and text—two inherently ambiguous modalities—for monocular depth estimation. Our framework is shown in the Fig. \ref{architecture}.

\section{Related Work}
\vspace{-1mm}
\subsection{Monocular Depth Estimation}
\vspace{-1mm}
In the domain of supervised monocular depth estimation using neural networks, the seminal work of \cite{eigen2014depth} constitutes a pivotal and foundational contribution. Their research introduces a coarse-to-fine convolutional neural network. On the basis of his model, a range of methods have emerged. It can be categorized into two distinct directions: one that involves the monodepth estimation problem as a pixel-wise regression task \cite{huynh2020guiding}, and another that formulates it as a pixel-wise classification challenge \cite{cao2017estimating}. In the decoding stage, \cite{lee2019big} introduced a local planar guidance layer to infer the plane coefficients, which were used to recover the full depth of resolution of the map. 
Self-supervised depth estimation from monocular videos or stereo image pairs is emerging due to the mitigation of manual labeling efforts. In the realm of monocular depth, \cite{zhou2017unsupervised} pioneered a self-supervised framework. This was achieved by jointly training depth and pose networks anchored on an image reconstruction loss. Subsequently, multiple frameworks \cite{godard2019digging, lu2023bird, lu2023deep} utilized a minimum re-projection loss and auto-masking loss. Based on sensor fusion method \cite{newcombe2011kinectfusion}, several studies \cite{guizilini20203d, chawla2021multimodal, lu2024slam} addressed the inherent scale ambiguity of monocular SfM-based methods through integration with other sensors. 

\vspace{-1mm}
\subsection{Geometric priors}
\vspace{-1mm}
Geometric priors have gained traction in monocular depth estimation, with the normal constraint \cite{long2021adaptive, qi2018geonet, lu2025} being particularly prevalent. This constraint enforces consistency between normal vectors derived from estimated depths and their ground truth counterparts. The piecewise planarity prior \cite{chauve2010robust, gallup2010piecewise, bodis2014fast} also provides a practical approximation for real-world scenes by segmenting them into 3D planes \cite{yang2018recovering, zhang2020geolayout, lu2025}, aiming to categorize the scene into primary depth planes. Despite inherent ambiguities in monocular depth estimation, most supervised learning paradigms still rely heavily on ground truth labels. Even as new architectures, such as Transformers, improve prediction accuracy, they do not address the core challenges of monocular depth estimation errors. While geometric priors help reduce some uncertainty, their overall impact remains limited. Departing from traditional geometric priors, we leverage camera model parameters to compute scene depth directly. Surface normal methods \cite{xue2020toward, wagstaff2021self} utilize camera parameters to calculate surface normals and obtain camera height, which is then used to compute the scale factor.
They primarily address how to use camera parameters for scale calculation. \cite{zhang2024embodiment,yang2023gedepth,cecille2024groco} discusses treating the camera model as depth prior information for depth estimation. Our method, which combines linguistic information and depth prior, can provide better depth predictions, thereby enhancing the result of the model.

\subsection{CLIP-based Depth Estimation}
Initially, \cite{zhang2022can} proposed a CLIP-based depth estimation model that generates prompt sentences to describe object distance. A text encoder extracts features from these prompts, which are then dot-multiplied with the image embodiment to compute depth weights. Depth estimation is generated through softmax and a linear combination of quantized depth bins. Building on DepthCLIP \cite{zhang2022can}, \cite{hu2024learning} train on a single image from each scene category to create a learnable depth codebook, storing quantized depth bins for each scene. 
\cite{zhang2025language} uses CLIP and depth information for image fusion.
\cite{auty2023learning} further improved DepthCLIP \cite{zhang2022can} by using continuous learnable tokens instead of discrete human language words. \cite{zeng2024wordepth} employs text descriptions corresponding to specific images as language priors, designing a variational autoencoder structure for depth estimation.
In our model, we embody vision and language into the system, combining Embodied Scene Depth with RGB image features to provide geometric and visual details for depth estimation. Text descriptions that include environmental depth information and semantic content are embodied as another depth prior, enhancing the model's ability to capture scene features and improving the accuracy of depth estimation.
\vspace{-1mm}
\section{Camera-Model  Embodied Depth 
}
\vspace{-1mm}
\label{sec:physics}
\subsection{Embodied Road Perception}

This study proposes a monocular depth estimation method that integrates the camera model into the system, enabling real-time interaction with the environment and dynamically updating depth priors according to changes in the road environment. This method leverages fundamental physical principles and comprehensively utilizes the intrinsic and extrinsic parameters of the camera as well as real-time semantic segmentation results to compute the absolute depth of road regions.
Initially, we project the camera's field of view onto a flat surface, calculating the depth for each pixel under the planar assumption. Given that roads typically satisfy the planar condition, we directly apply off-the-shelf semantic segmentation model to identify road regions within the image, extracting these areas from the planar depth map as the absolute depth for roads, defined as "Embodied Road Depth." For adjacent ground areas that are nearly flat, their actual depth closely matches the computed road depth; thus, we retain the depth values of these regions, defining them as "Embodied Plane Depth." Additionally, based on common scene configurations, we assume that the depth of pedestrians, vehicles, or trees on the road aligns with the depth of the road beneath them. Consequently, we extend the planar depth to these vertical surfaces, defining this as "Extended Embodied Plane Depth." Lastly, to address gaps in the scene's depth information, we employ image inpainting techniques \cite{telea2004image} to obtain a "Embodied Scene Depth."

\begin{figure}[t]
\begin{center}
\includegraphics[width=8cm, height=5cm]{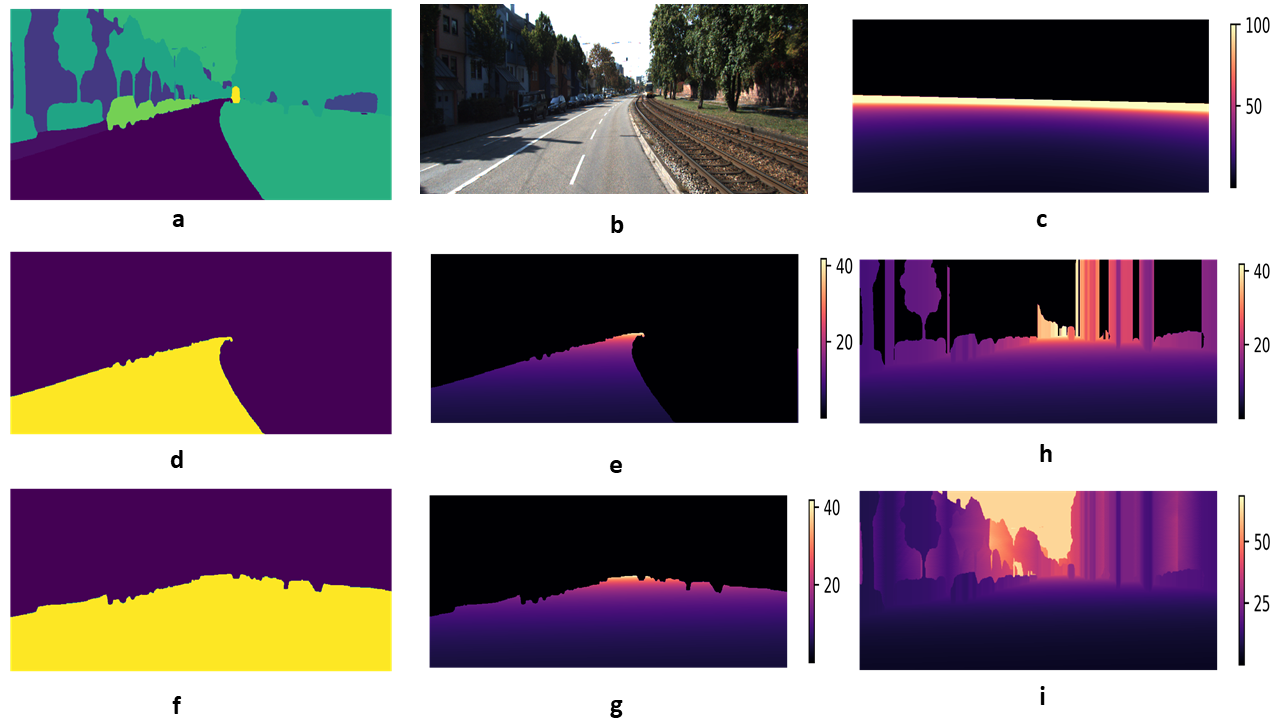}
\end{center}
\vspace{-6mm}
\caption{Embodied Depth Perception on KITTI: (a) Semantic segmented image; (b) RGB image; (c) Embodied Surface Depth; (d) Road segmented from semantic segmented image; (e)Embodied Road Depth; (f) Ground segmented from semantic segmented image; (g) Embodied Ground Depth (h) Extended Embodied Ground Depth; (i) Embodied Scene Depth. }
\label{Depth Methodology}
\vspace{-5mm}
\end{figure}

We calculate the depth of each pixel in the image under a planar assumption.
The transformation of a three-dimensional point from the world coordinate system \((x_w, y_w, z_w)\) to the camera coordinate system \((x_c, y_c, z_c)\), followed by its projection from the camera coordinate system to the two-dimensional image plane \((u, v)\), can be accurately described using the following linear camera model:
\vspace{-3mm}
\begin{equation}
\vspace{-1.52mm}
\centering
\label{backproject}
\begin{array}{l}
{z_c}\left[ {\begin{array}{*{20}{c}}
{x'}\\
{y'}\\
1
\end{array}} \right] = \left[ {\begin{array}{*{20}{c}}
K&0
\end{array}} \right]\left[ {\begin{array}{*{20}{c}}
R&T\\
0&1
\end{array}} \right]\left[ {\begin{array}{*{20}{c}}
{{X_w}}\\
{{Y_w}}\\
{{Z_w}}\\
1
\end{array}} \right]
\end{array}
\end{equation}
In this context, \( \mathbf{K} \) denotes the camera's intrinsic matrix, while \( \mathbf{R} \) represents the rotation matrix, and \( \mathbf{T} \) is the translation vector, collectively comprising the camera's extrinsic parameters. Substituting \( \mathbf{K} \), \( \mathbf{R} \), and \( \mathbf{T} \) into Eq. \ref{backproject} yields the following equation:
\vspace{-2mm}
\begin{equation}
\small
\setlength{\arraycolsep}{2pt} 
{Z_c}\left[ \begin{array}{c} \label{backprojectdetail}
u\\
v\\
1
\end{array} \right] = \left[ \begin{array}{ccc}
{{f_x}} & 0 & {{O_x}}\\
0 & {{f_y}} & {{O_y}}\\
0 & 0 & 1
\end{array} \right]\!\left[ \begin{array}{cccc}
{{x_{11}}} & {{x_{12}}} & {{x_{13}}} & {{t_x}}\\
{{x_{21}}} & {{x_{22}}} & {{x_{23}}} & {{t_y}}\\
{{x_{31}}} & {{x_{32}}} & {{x_{33}}} & {{t_z}}\\
0 & 0 & 0 & 1
\end{array} \right]\!\left[ \begin{array}{c}
{{x_w}}\\
{{y_w}}\\
{{z_w}}\\
1
\end{array} \right]
\end{equation}

In this framework, \( u, v \) represent the pixel coordinates on the image plane, where the origin of the coordinate system is located at the optical center \((O_x, O_y)\) of the image, often referred to as the principal point. The terms \( f_x, f_y \) denote the camera's focal lengths along the x and y axes, respectively.
The combined transformation of the camera's intrinsic and extrinsic matrices can be expressed as A, defined as follows:
\vspace{-4mm}
\begin{equation}
\centering
\begin{array}{l}
\small
\setlength{\arraycolsep}{2pt}
A = \left[ {\begin{array}{*{20}{c}}
K&0
\end{array}} \right]\left[ {\begin{array}{*{20}{c}}
R&T\\
0&1
\end{array}} \right] = \left[ {\begin{array}{*{20}{c}}
{\begin{array}{*{20}{c}}
{{a_{11}}}&{{a_{12}}}&{{a_{13}}}\\
{{a_{21}}}&{{a_{22}}}&{{a_{23}}}\\
{{a_{31}}}&{{a_{32}}}&{{a_{33}}}
\end{array}}&{\begin{array}{*{20}{c}}
{{a_{14}}}\\
{{a_{24}}}\\
{{a_{34}}}
\end{array}}
\end{array}} \right]
\end{array}
\vspace{-3mm}
\end{equation}
By substituting A in Eq. \ref{backprojectdetail},
\vspace{-2mm}
\begin{equation}
\centering
\begin{array}{l}
{z_c}u = {a_{11}}{x_w} + {a_{12}}{y_w} + {a_{13}}{z_w} + {a_{14}}\\  \label{zcuzcv}
{z_c}v = {a_{21}}{x_w} + {a_{22}}{y_w} + {a_{23}}{z_w} + {a_{24}}\\
1 = {a_{31}}{x_w} + {a_{32}}{y_w} + {a_{33}}{z_w} + {a_{34}}
\end{array}
\vspace{-2mm}
\end{equation}

If we know that the height of the world coordinate system above the ground is \( h \), and assuming that the Y-axis of the world coordinate system points towards the ground, then \( y_w = h \). By Substituting \( y_w = h \) in Eq. \ref{zcuzcv}, we solve for \( x_w\), \( z_w\), and \( z_c\). Utilizing the world coordinate system \((x_w, y_w, z_w)\), we can derive the precise camera coordinates \((x_c, y_c, z_c)\) using the following equation:
\vspace{-3mm}
\begin{equation}
\centering
\begin{array}{l}
\left[ {\begin{array}{*{20}{c}} \label{world_to_cameraplane}
{{x_c}}\\
{{y_c}}\\
{{z_c}}
\end{array}} \right] = \left[ {\begin{array}{*{20}{c}}
R&T\\
0&1
\end{array}} \right]\left[ {\begin{array}{*{20}{c}}
{{x_w}}\\
{{y_w}}\\
{{z_w}}\\
1
\end{array}} \right]
\end{array}
\vspace{-2mm}
\end{equation}
From Eq. \ref{world_to_cameraplane}, we can derive camera coordinates \((x_c, y_c, z_c)\) for pixel \((u, v)\).
Also by substituting Eq. \ref{world_to_cameraplane} in Eq. \ref{backproject}, we get
\vspace{-2mm}
\begin{equation}
\small 
\setlength{\arraycolsep}{2pt} 
{Z_c}\left[ \begin{array}{c} \label{camera_to_imageplane}
u\\
v\\
1
\end{array} \right] = \left[ \begin{array}{ccc}
{{f_x}} & 0 & {{O_x}}\\
0 & {{f_y}} & {{O_y}}\\
0 & 0 & 1
\end{array} \right]\!\left[ \begin{array}{c}
{{x_c}}\\
{{y_c}}\\
{{z_c}}\\

\end{array} \right]
\end{equation}

From Eq. \ref{camera_to_imageplane}, if the height of the camera in the camera coordinate system, denoted as \( y_c = h \), is known, we can directly solve for \( x_c \) and \( z_c \). Utilizing the coordinates \((x_c, y_c, z_c)\), we generate a 3D point cloud representing a flat surface and calculate the point-to-point distance from the camera's center, (0,0,0), to a point on the ground.
Assuming the entire camera field of view is planar, we calculate absolute depths for each pixel. We then identify actual planar regions using semantic segmentation to isolate areas corresponding to the plane, such as roads in the KITTI dataset, creating Embodied Road
Depth. Our method has been evaluated on the  KITTI \cite {geiger2013vision} dataset.

\subsection{Embodied Scene Depth}
\vspace{-1mm}
In this paper, we elaborate on a real-time embodied depth perception method based on camera model embodying, which has been validated on the KITTI dataset and demonstrates astonishing accuracy in road areas. Compared to the sparse ground truth, this method provides denser depth. However, this specificity may lead to overfitting to road regions during model training, potentially limiting its applicability in diverse environments. To mitigate this risk and enhance the model’s adaptability, we extended the application scope of embodied depth perception to encompass the entire image scene. 
Specifically, we first generalize road depth to all flat surfaces within the camera’s view (such as roads, sidewalks, parking lots), as these areas often exhibit uniform flatness. Additionally, according to real-world principles, the depth of vertical objects (such as vehicles, pedestrians, buildings) aligns with the depth of the ground they are connected to. By extending depth calculations from flat regions to vertical surfaces and propagating depth upward along the boundaries between these surfaces, we generate a comprehensive scene depth representation—Embodied Scene Depth, which can interact with the scene in real-time to compute scene depth priors.

After the vertical extension of the physical depth, some pixels may still lack depth information. We use the Telea Inpainting Technique \cite{telea2004image} to fill in these missing parts, leveraging the existing depth embedding. The Telea inpainting method was chosen for its efficiency in incorporating the directional rate of change and geometric distance of neighboring pixels, as well as its faster processing speed compared to complex deep learning-based inpainting methods, proving effective in these situations. For objects not in contact with the ground, we extend the depth from intermediary objects to the ground. The sky region is filled as infinity, ultimately generating a dense, gap-free Embodied Scene Depth. This step is crucial for creating depth priors that significantly enhance depth estimation results. Our method has been validated on the KITTI \cite{geiger2013vision} dataset, with the results showing relatively high accuracy compared to the ground truth, especially on flat surfaces.

 we categorized five distinct types of Embodied Depth Perception: Embodied Surface Depth, Embodied Road Depth, Embodied Ground Depth, Extended Embodied Ground Depth, and Embodied Scene Depth. Using the KITTI dataset, we present visual examples of these different categories in Fig \ref{Depth Methodology}. Initially, a pre-trained segmentation model is applied to obtain the segmentation result (a), where (d) and (f) specifically denote road and flat regions. Following this, images (c), (e), (h), (g), and (i) serve as visual representations for each stage of embodied depth perception, corresponding to various types of embodied depth. These visualizations provide a deeper understanding of the unique contributions of each embodied depth category to the overall depth estimation process.

\section{Language-Vision Embodied Depth}
Depth estimation aims to infer the most probable scene from all possible distributions based on image conditions. We introduce language priors, which are semantically related and provide scale information. While images reveal layout and object shapes, text offers a strong prior on scale. Combining the complementary strengths of both modalities enables more accurate depth estimation.
\subsection{Depth-Guided Text Variational Auto-Encoder}

Using language to describe depth information provides prior knowledge and semantic guidance for depth estimation, constraining the solution space by expressing spatial relationships and scene layouts. We design a variational auto-encoder (VAE) that leverages a pre-trained CLIP text encoder to extract feature embodiment from input descriptions, building a shared latent space to capture contextual and spatial details in the text. The embodiments are then input into a multi-layer perception (MLP) to estimate the mean \(\hat{\mu} \in \mathbb{R}^d\) and standard deviation \(\hat{\sigma} \in \mathbb{R}^d\) of the latent distribution, modeling the probabilistic distribution of plausible scene layouts, to generate depth consistent with the characteristics depicted in the text.

\textbf{Text description.}
For textual descriptions, we use ExpansionNet-v2 \cite{hu2022expansionnet} to extract captions from the images. For depth descriptions, we obtain each object's depth and distance ranking based on its embodiment scene depth and semantic segmentation results, and convert this information into depth descriptions. Specifically, let the set of objects in the image be \(\{O_1, O_2, \dots, O_N\}\), where \(d_i\) represents the embodiment scene depth of object \(O_i\). Based on the depth  \(\{d_1, d_2, \dots, d_N\}\), we can define the relative ranking \(r_i\) for each object: $r_i = \text{rank}(d_i), $ 
\(\text{rank}(d_i)\) denotes the position of the \(i\)-th object in a sorted list of all objects by depth from nearest to farthest.
The textual description for each object \(O_i\) can be represented as:
$T_i$ = “This object seems to be $d_i$  and ranks as the  $r_i$-th farthest in distance.”
We combine the textual descriptions and depth information into a single representation, \( T_{\text{combined}} = \{t_1, t_2, \dots, t_n; \, \text{t}_{d1}, \text{t}_{d2}, \dots, \text{t}_{dn} \} \), where each \( t_i \) represents a specific semantic description extracted from the image, and each \( \text{t}_{di} \) (depth description) contains depth information and the relative position of the object. This combined text serves as the input for the text encoder.
To encode the textual description \( t \), we first utilize the CLIP text encoder to obtain a high-dimensional feature embodying. This embodiment is processed by a multi-layer perceptron (MLP) to estimate the mean and standard deviation of the latent distribution, represented as \((\hat{\mu}, \hat{\sigma}) = g(t) \in \mathbb{R}^{2 \times d}\), where \(\hat{\mu}\) and \(\hat{\sigma}\) are the mean and standard deviation vectors of the latent distribution, modeling the plausible scene layout’s latent space. To sample from this distribution, we apply the reparameterization trick to ensure the sampling process is differentiable, enabling gradient backpropagation for optimization.
Specifically, the reparameterization trick introduces a noise variable \(\epsilon \sim \mathcal{N}(0, 1)\) drawn from a standard Gaussian distribution, transforming the sampling process into a differentiable form. We sample from the latent distribution using the following equation:
\vspace{-1mm}
\begin{equation}
    \hat{z} = \hat{\mu} + \epsilon \cdot \hat{\sigma}
    \vspace{-1mm}
\end{equation}
\noindent where \(\hat{z}\) is the latent variable, combining the prior information from the textual description with stochasticity. 
We use the ResNet-50 of the CLIP\cite{zamir2022restormer} text encoder to extract text features, setting the latent space dimension d of both the text-embodying and the image conditional sampler to 128. When the text features extracted by CLIP are of dimension 1024, we use a 3-layer MLP with hidden dimensions of 512, 256, and 128 to encode them.
We employ the KL divergence loss as a regularization method to pull the predicted latent distribution (parameterized by the mean \(\mu\) and standard deviation \(\sigma\)) toward a standard Gaussian distribution. The KL divergence loss applied to \(\mu\) and \(\sigma\) is defined as:
\vspace{-1mm}
\begin{equation}
  \mathcal{L}_{\text{KL}}(\mu, \sigma) = -\log(\sigma) + \frac{\sigma^2 + \mu^2}{2} - \frac{1}{2}.  
\end{equation}

To enhance the model's robustness to scale variations, we employ the Scale-Invariant Logarithmic Loss (SiLog Loss), which performs especially well in scenarios with different scales. The SiLog Loss is defined as:

\vspace{-1mm}
\begin{scriptsize}
\begin{equation}
   \mathcal{L}_{\text{SiLog}} = \frac{1}{n} \sum_{i=1}^n \left( \log(y_i) - \log(\hat{y}_i) \right)^2 - \frac{1}{n^2} \left( \sum_{i=1}^n \left( \log(y_i) - \log(\hat{y}_i) \right) \right)^2 
\end{equation}
\end{scriptsize}
\noindent where \( y_i \) denotes the ground-truth,
\( \hat{y}_i \) denotes the predicted depth, \( n \) is the number of pixels in image.

\subsection{Embodied-Driven Feature Fusion and Conditional Sampling}
We take the depth distribution generated by the text-VAE as the latent prior distribution for the corresponding image, and by sampling the latent variable, we obtain a depth based on the text prior. To achieve this, we introduce an image-based conditional sampler that performs sampling under image conditions. This sampler predicts a noise vector \(\tilde{\epsilon}\), replacing \(\epsilon \sim N(0,1)\) drawn from a standard Gaussian distribution, and uses this noise vector to generate the latent variable \(\tilde{z} = \hat{\mu} + \tilde{\epsilon} \cdot \hat{\sigma}\), which is then decoded by the depth decoder to produce the depth.
To ensure that the generated depth not only aligns with the text prior distribution but also accurately reflects the actual scene structure in the image, we integrate embodiment scene depth and RGB image features. embodiment scene depth provides a reliable geometric prior, guiding the sampling process to focus on depth ranges consistent with physical constraints, preventing unrealistic depth values. Meanwhile, RGB image features add visual information such as texture, color, and shape. By using cross-attention to merge these two types of features, the sampled latent variable meets both geometric constraints and captures visual details, thereby enhancing the accuracy of depth estimation.

\textbf{Image Encoder.} The RGB image and physics depth serve as the inputs to the encoder. To extract both image and depth information, we use Transformer/Restormer \cite{zamir2022restormer}-based blocks as the primary feature extractor, as shown in the following formula.
\vspace{-2mm}
\begin{equation}
     F_{rgb} = \mathcal{F}^{I}_{r}(I_{rgb}), F_{dep} = \mathcal{F}^{I}_{d}(I_{dep})
     \vspace{-2mm}
\end{equation}

\noindent where $I_{rgb} \in \mathbb{R}^{H \times W \times 3}$ and $I_{dep} \in \mathbb{R}^{H \times W \times 1}$ represent the images and physcis depth. $H, W$ denote the height and width of the image. $\mathcal{F}^{I}_{r}$ and $\mathcal{F}^{I}_{d}$ are the  image and  depth encoders, respectively.

\textbf{Cross-Modality Feature Fusion.}
The depth provides spatial and geometric information, while the RGB image captures texture and color details. We use a cross-attention mechanism in the cross-modality fusion layer to integrate features from both modalities, enhancing the interaction between depth and RGB features and enabling efficient information transfer. The fused features retain the geometric structure from the depth information and incorporate the details from the RGB image.
\vspace{-2mm}
\begin{equation}
\{Q_{r}, K_{r}, V_{r}\} = \mathcal{F}^{r}_{qkv}(F_{rgb})
   \vspace{-2mm} 
\end{equation}
\begin{equation}
\{Q_{d}, K_{d}, V_{d}\}= \mathcal{F}^{d}_{qkv}(F_{dep})
   \vspace{-2mm} 
\end{equation}
\noindent where $F_{rgb},F_{dep}$ denote features from the image and depth encoder. Subsequently, we exchange the queries $Q$ of two modalities for spatial interaction:
\vspace{-2mm}
\begin{equation}
    F_{f}^{d}=softmax(\frac{Q_{r}K_{d}}{d_{k}})V_{d}, 
    \vspace{-2mm}
\end{equation}
\begin{equation}
 F_{f}^{r}=softmax(\frac{Q_{d}K_{r}}{d_{k}})V_{r}  
 \vspace{-2mm}
\end{equation}
\noindent where $d_{k}$ is the scaling factor. Finally, we concatenate the results obtained by the cross-attention through $F_{f}^{0}=Concat({F_{f}^{d},F_{f}^{r}})$ to get the fusion features.

\textbf{Conditional Sampler.} 
We obtain the distribution of possible 3D scene layouts through language embodiment, allowing us to generate depth maps from textual features via the depth decoder. However, the depth predicted from text cannot correspond pixel-by-pixel with the image. The mean and variance learned by the text-VAE represent the latent prior distribution of possible scene layouts. To predict the depth map \(\tilde{y}\) for an image \(x\), we need to sample from the latent distribution corresponding to the 3D scene layout, leveraging an image-based conditional prior. For this purpose, we introduce a conditional sampler based on the fusion of embodied scene depth and image features, which predicts a sample \(\tilde{\epsilon}\) to replace \(\epsilon \sim N(0,1)\) from the standard Gaussian distribution. As before, using the reparameterization trick, we use \(\tilde{\epsilon}\) to select the latent vector \(\tilde{z}\), which is then decoded by the depth decoder to produce the depth.

Based on the embodiment scene depth and image features, the Transformer-based blocks encode downsampled features into \( h \times w \) local samples \(\tilde{\epsilon} \in \mathbb{R}^{d \times h \times w}\), which are then used to sample from the latent distribution of our textual description. The mean \(\hat{\mu}\) and standard deviation \(\hat{\sigma}\) also serve as inputs to this process. We refer to this module as the conditional sampler, \(\tilde{\epsilon} = f'(x, \hat{\mu}, \hat{\sigma})\), which aims to estimate the most probable latent variable for the text-VAE. Consequently, the latent vector for scene layout is expressed as 
\(\tilde{z} = \hat{\mu} + \tilde{\epsilon} \cdot \hat{\sigma}\), with the predicted depth given by \(\tilde{y} = h(\tilde{z})\).
Our depth decoder uses upsampling Transformer-based blocks and shares weights with the depth decoder in the textual module. 

\textbf{Training: } Following \cite{zeng2024wordepth}, we use an alternating optimization scheme to jointly train the text-VAE with the conditional sampler. 
In one alternating step, we freeze the conditional sampler and train the text-VAE and depth decoder. This involves predicting \(\hat{\mu}\) and \(\hat{\sigma}\) from the text description \(t\), and using the reparameterization trick with \(\epsilon\) sampled from a standard Gaussian distribution to generate the latent vector. 
In the next alternating step, we freeze the text-VAE and train the conditional sampler with the depth decoder. Here, we use the frozen text-VAE to predict \(\hat{\mu}\) and \(\hat{\sigma}\), and sample from the latent distribution using \(\tilde{\epsilon}\) predicted from the image. 
The alternating steps are repeated in a ratio of \(p\) (text-VAE) to \(1 - p\) (conditional sampler).

\vspace{-1mm}
\section{Experiments}
\label{exper}
\vspace{-1mm}
\subsection{Datasets}

We conducted a comprehensive evaluation of our proposed method on two depth estimation datasets: KITTI \cite{geiger2013vision} and Dense Depth for Autonomous Driving (DDAD) \cite{guizilini20203d}. We followed the Eigen split \cite{eigen2014depth}, which consists of 23,158 training images and 697 test images. To ensure consistency, we adopted the cropping method defined in \cite{garg2016unsupervised} and upsampled the predicted depth to match the resolution of the ground truth for a more precise comparative analysis.
In contrast, DDAD serves as a more challenging benchmark for depth estimation, featuring diverse geographic environments, multiple camera perspectives, and an extended depth prediction range. This dataset comprises 150 scenes (12,650 training images per camera) and 50 scenes (3,950 test images per camera). We adhered to the standard evaluation protocol defined in \cite{guizilini2021sparse}, utilizing four key camera views: forward, backward, left forward, and right forward.




\subsection {Embodied Depth Perception}
\label{section:Physics Depth Demonstration}

\begin{figure}[t]
\begin{center}
\includegraphics[width=8cm, height=3cm]{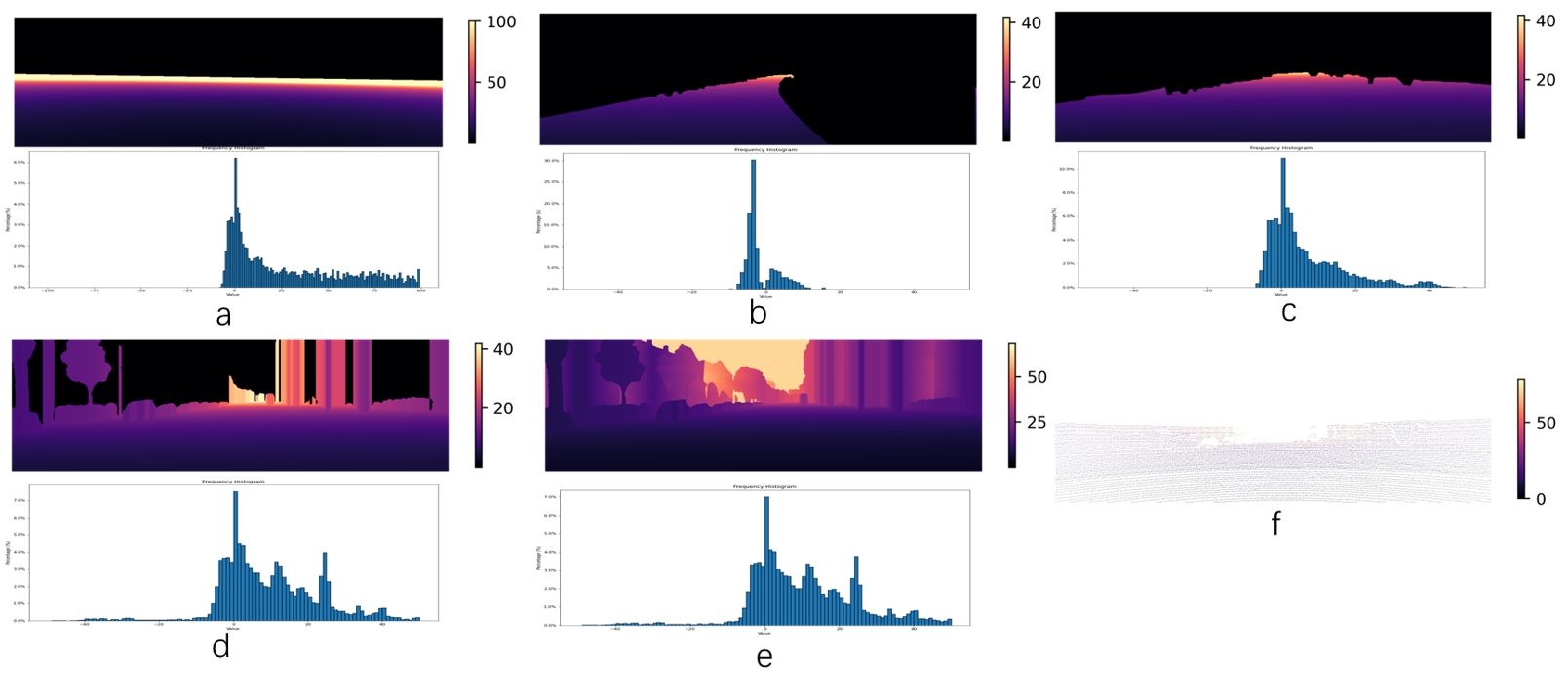}
\end{center}
\vspace{-8mm}
\caption{Error distribution of Embodied Depth: (a) Embodied Surface Depth and error distribution; (b) Embodied Road Depth and error distribution; (c) Embodied Ground Depth and error distribution; (d) Extended Embodied Ground Depth and error distribution; (e) Embodied Scene Depth and error distribution; (f) Sparse LiDAR depth  as Ground Truth.  }
\label{Physics Depth Ablation}
\vspace{-2mm}
\end{figure}

\textbf{Error distribution:} In our comparative analysis, as depicted in Fig. \ref{Physics Depth Ablation} and Table \ref{image error}, we examine the efficacy of the embodied depth perception on a sample image. The results clearly indicate a high proficiency in estimating road surfaces (b), which are nearly flat. This is substantiated by over 99\% of pixels demonstrating less than 10\% error and more than 81\% of pixels displaying less than 5\% error compared to Ground Truth. 
These findings reinforce our assertion that for road surfaces, or any comparably flat terrains, our embodied depth perception can effectively replace LiDAR . 
However, as expected, the precision of our embodied depth estimation diminishes when incorporating more complex surfaces such as sidewalks, rail tracks, parking lots, and ground. This decrease in accuracy is attributed to the non-uniformity in their level relative to the camera base, coupled with the uncertainty of their flatness. Additionally, the application of embodied depth perception to vertical surfaces incurs a minor increase in error. Despite this, the resultant trade-off is deemed a necessary compromise for attaining a more comprehensive, denser depth, thereby enhancing the overall depth estimation. 

\begin{table}[t]
\begin{center}
\resizebox{0.47\textwidth}{!}{
\begin{tabular}{llllll}
\hline
\ & \cellcolor{pink}Embodied  & \cellcolor{pink}Embodied & \cellcolor{pink}Embodied & \cellcolor{blue!30}Extended  & \cellcolor{blue!30}Embodied \\
 & \cellcolor{pink}Surface  & \cellcolor{pink}Road  & \cellcolor{pink}Ground  & \cellcolor{blue!30}Embodied & \cellcolor{blue!30}Scene \\ 
  & \cellcolor{pink} Depth & \cellcolor{pink} Depth & \cellcolor{pink} Depth & \cellcolor{blue!30}Ground Depth & \cellcolor{blue!30} Depth \\ 
\hline
+/- 5\% error  & 47.29\%  & 80.24\%  & 60.30\%  & 41.83\%  & 38.88\%  \\
\hline
+/- 10 \% error  & 58.34\%  & 99.33\%  & 74.89\%  & 55.44\%  & 52.45\%  \\
\hline
\end{tabular}}
\end{center}
\vspace{-7mm}
\caption{Embodied depth perception in a sample KITTI image. }
\label{image error}
\vspace{-2mm}
\end{table}

\begin{figure}[t]
\begin{center}
\includegraphics[width=8.2cm, height=5cm]{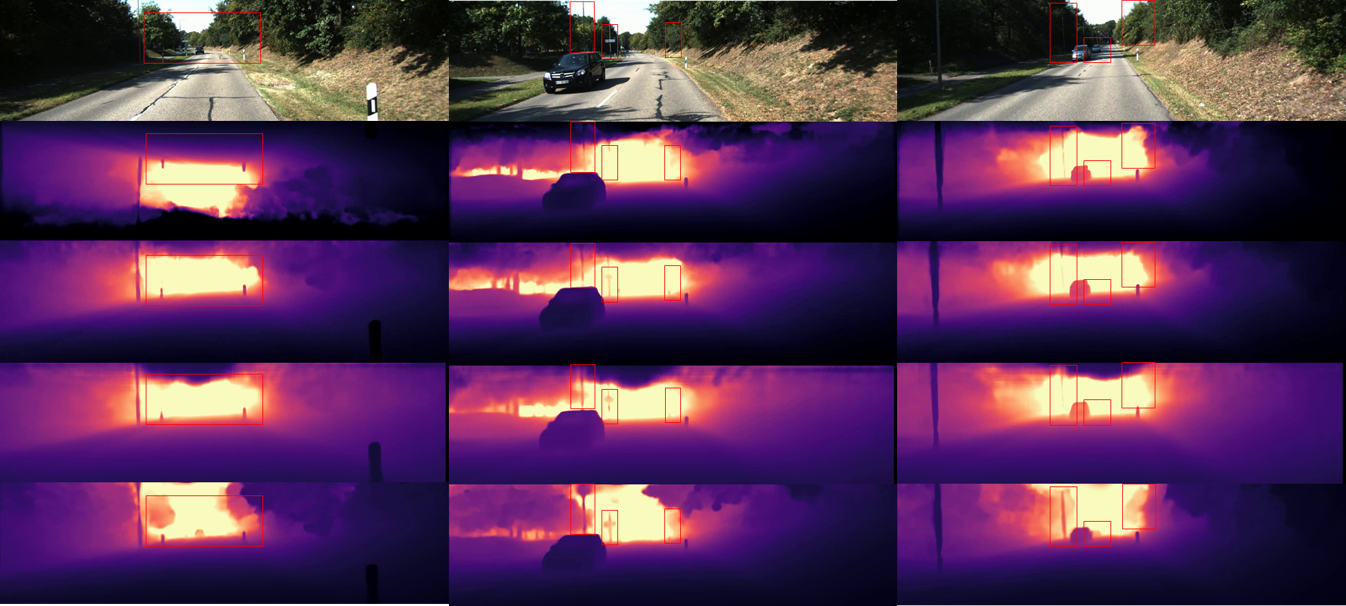}
\end{center}
\vspace{-6mm}
\caption{\textbf{Visual results on KITTI \cite{geiger2013vision}:} From top to bottom; the models are ECoDepth \cite {Patni_2024_CVPR}; MIM \cite{xie2023revealing}; AFNet \cite{cheng2024adaptive} ; Ours. }
\vspace{-4mm}
\label{qualitative}
\end{figure}

\textbf{Error of Semantic Segmentation:} In calculating embodied depth, we begin with semantic segmentation to locate the ground within the image and compute its depth. Given the current maturity of Semantic segmentation models, we utilize multiple pre-trained models for image segmentation. As shown in Table \ref{segment error}, the variations in embodied depth resulting from different Semantic segmentation models are remarkably similar, even when compared with the semantic segmentation ground truth from the KITTI dataset. This consistency indicates that our depth calculation accuracy has low dependency on the segmentation model’s performance, as differences among models have minimal impact on ground depth accuracy. Our model demonstrates low reliance on segmentation models when calculating and extending ground depth.

\begin{table}[t]
\begin{center}
\vspace{-2mm}
\resizebox{0.47\textwidth}{!}{
\begin{tabular}{lllll}
\toprule
KITTI Date  & 
\cellcolor{pink}\begin{tabular}[c]{@{}l@{}}{Embodied Road} \\{Depth Error:} \\ {+/- 5\%}\end{tabular} & 
\cellcolor{pink}\begin{tabular}[c]{@{}l@{}}{Embodied Road}\\{Depth Error:} \\ {+/- 10\%}\end{tabular} & 
\cellcolor{blue!30}\begin{tabular}[c]{@{}l@{}}{Embodied Ground} \\{Depth Error:} \\ { +/- 5\%}\end{tabular} & 
\cellcolor{blue!30}\begin{tabular}[c]{@{}l@{}}{Embodied Ground} \\{Depth Error:} \\ { +/- 10\%}\end{tabular} \\
\midrule
DeeperLab \cite{yang2019deeperlab} &82.97\% &93.89\% &76.74\% &86.93\%\\
AUNet \cite{li2019attention} &83.12\% &93.91\% &77.05\% &87.14\%\\
Panoptic-DeepLab \cite{cheng2020panoptic} &83.21\% &94.05\% &77.17\% &87.24\%\\
UPSNet \cite{xiong2019upsnet} &83.24\% &94.11\% &72.19\% &87.29\%\\
EfficientPS \cite{mohan2021efficientps} &83.31\% &94.15\% &72.27\% &87.31\%\\
KITTI Ground Truth \cite{geiger2013vision} &83.75\% &94.36\% &72.48\% &87.82\%\\

\bottomrule
\end{tabular}}
\end{center}
\vspace{-6mm}
\caption{Comparison of Embodied Depth Errors on the KITTI Dataset under different semantic  segmentation}
\label{segment error}
\vspace{-2mm}
\end{table}


\vspace{-1mm}
\subsection {Evaluation of Embodied Depth Perception}
In this study, we have produced embodied depth for the entire KITTI dataset, a crucial step in training our models. This process entailed a detailed examination of the variances between embodied road depth, embodied ground depth within these datasets. As delineated in Tables \ref{kitti error} , the KITTI dataset revealed that approximately 90\% of pixels had an error margin of less than 10\%, and around 80\% of pixels maintained a deviation of less than 5\% compared to Ground Truth. 
The data, as presented in Tables \ref{kitti error}, suggest that the embodied road depth surpasses the embodied ground depth in terms of accuracy. However, the number of road pixels in a single image is limited. To augment the density of embodied depth pixels in each image, we extended the application of embodied road depth to embodied ground depth, albeit recognizing the slight variations in their flatness. This expansion, while increasing the number of data points, concurrently broadens the error margin compared to the ground truth. Nonetheless, embodied ground depth maintains commendable accuracy despite elevated error levels, thereby enriching the dataset and diminishing the risk of model overfitting.


\begin{table}[t]
\begin{center}
\resizebox{0.47\textwidth}{!}{
\begin{tabular}{lllll}
\toprule
KITTI Date  & 
\cellcolor{pink}\begin{tabular}[c]{@{}l@{}}{Embodied Road} \\{Depth Error:} \\ {+/- 5\%}\end{tabular} & 
\cellcolor{pink}\begin{tabular}[c]{@{}l@{}}{Embodied Road}\\{Depth Error:} \\ {+/- 10\%}\end{tabular} & 
\cellcolor{blue!30}\begin{tabular}[c]{@{}l@{}}{Embodied Ground} \\{Depth Error::} \\ { +/- 5\%}\end{tabular} & 
\cellcolor{blue!30}\begin{tabular}[c]{@{}l@{}}{Embodied Ground} \\{Depth Error::} \\ { +/- 10\%}\end{tabular} \\
\midrule
2011-09-26 &84.28\% &96.26\% &75.08\% &89\%\\
2011-09-28 &80.61\% &85.64\% &61.21\% &77\%\\
2011-09-29 &90.53\% &97.34\% &74.46\% &91\%\\
2011-09-30 &76.43\% &91.86\% &56.98\% &81\%\\
2011-10-03  &78.12\% &94.61\% &62.77\% &85\%\\

\bottomrule
\end{tabular}}
\end{center}
\vspace{-6mm}
\caption{Error between  embodied  depth and KITTI ground truth. The proportion of the 5-days  embodied road depth error and embodied ground depth error within $5\%$ and within $10\%$ of ground truth, respectively, in the KITTI dataset \cite{geiger2013vision}. }
\label{kitti error}
\vspace{-4mm}
\end{table}

\begin{table}[t]
\begin{center}
\resizebox{0.47\textwidth}{!}{
\begin{tabular}{lllllllll}
\multicolumn{1}{l}{\bf\ Method}  & \multicolumn{1}{l}{\cellcolor{pink}AbsRel ↓} & \multicolumn{1}{l}{\cellcolor{pink}Sq Rel↓} & \multicolumn{1}{l}{\cellcolor{pink}RMSE↓} & \multicolumn{1}{l}{\cellcolor{pink}RMSE log↓} & \multicolumn{1}{l}{\bf\cellcolor{blue!30}$\delta < 1.25$\ ↑} & \multicolumn{1}{l}{\bf\cellcolor{blue!30}$\delta < 1.25^2$\ ↑} & \multicolumn{1}{l}{\bf\cellcolor{blue!30}$\delta < 1.25^3$\ ↑}
\\ \hline 
Eigen \cite{eigen2015predicting}   &0.203 &1.548 &6.307 &0.282 &0.702 &0.898 &0.967
\\ \hline 
DORN \cite{fu2018deep}  &0.071 &0.268 &2.271 &0.116 &0.936 &0.985 &0.995
\\  \hline 
VNL \cite{yin2019enforcing}      &0.072 &-     &3.258 &0.117 &0.938 &0.990 &0.998
\\  \hline 
BTS \cite{lee2019big}       &0.061 &0.261 &2.834 &0.099 &0.954 &0.992 &0.998
\\  \hline 
Adabins \cite{bhat2021adabins}  &0.058 &0.190 &2.360 &0.088 &0.964 &0.995 &0.999
\\  \hline 
P3Depth \cite{patil2022p3depth}  &0.071 &0.270 &2.842 &0.103 &0.953 &0.993 &0.998
\\  \hline 
DepthFormer \cite{li2023depthformer}   &0.052 &0.158 &2.143 &0.079 &0.975 &0.997 &0.999
\\  \hline 
NeWCRFs \cite{yuan2022new}  &0.052 &0.155 &2.129 &0.079 &0.974 &0.997 &0.999
\\  \hline 
iDisc \cite{piccinelli2023idisc} &0.050	&0.145 &2.067 &0.077 &0.977	&0.997	&0.999
\\  \hline 
URCDC \cite{shao2023urcdc}   &0.050 &0.142 &2.032 &0.076 &0.977 &0.997 &0.999
\\  \hline 
Wordepth \cite{zeng2024wordepth} &0.049 &- &2.039&0.074 &0.979 &0.998& 0.999
\\  \hline

ECoDepth \cite{Patni_2024_CVPR} &0.048	&	0.139 &1.966	&0.074	&0.979	&0.998	&1.000	
\\  \hline  
MiM(large) \cite{xie2023revealing}      &0.050	 &0.139 &1.966	&0.075	&0.977	&0.998	&0.999	
\\  \hline  
Trap Attention \cite{ning2023trap}  &0.050 &0.128 &1.869 &0.074 &0.980 &0.998 &0.999
\\  \hline 
Depth Anything \cite{yang2024depth} &0.046&0.121 & 1.869 &0.069 &  0.982 &0.998 &1.000
 \\  \hline 
Metric3D v2 \cite{hu2024metric3d} &0.039&- &1.766 &0.060 &0.989 &0.998 &1.000 
 \\  \hline
UniDepth \cite{piccinelli2024unidepth}  &0.042	&-&1.75	&0.064	&0.986	&0.998	&0.999	
 \\  \hline
LightedDepth \cite{zhu2023lighteddepth} &0.041	&0.107 &1.748	&0.059	&0.989	&0.998&	0.999	
 \\  \hline

AFNet \cite{cheng2024adaptive}  &0.044 &0.132 &1.712	&0.069&	0.980	&0.997	&0.999
  \\  \hline
Ours   &0.0251   & 0.0428      &1.654 &    0.048   &0.991   &0.998   &0.999 
\\  \hline  
\end{tabular}}
\end{center}
\vspace{-7mm}
\caption{ For a quantitative depth comparison using the Eigen split of the KITTI dataset \cite{geiger2013vision}.}

\label{Kitti_result}
\vspace{-2mm}
\end{table}

\begin{table}[t]
\begin{center}
\resizebox{0.47\textwidth}{!}{
\begin{tabular}{ccccccc}
\multicolumn{1}{c}{\bf\ Method}   &\multicolumn{1}{c}{\cellcolor{pink}AbsRel ↓}   &\multicolumn{1}{c}{\cellcolor{pink}RMSE↓} &\multicolumn{1}{c}{\bf\cellcolor{blue!30}$\delta < 1.25$\ ↑}  &\multicolumn{1}{c}{\bf\cellcolor{blue!30}$\delta < 1.25^2$\ ↑}  &\multicolumn{1}{c}{\bf\cellcolor{blue!30}$\delta < 1.25^3$\ ↑} 

\\ \hline 
PackNet-SAN\cite{guizilini2021sparse}  & 0.187	& 11.936 	& 0.684 & 0.821 & 0.932
\\  \hline 
 BTS \cite{lee2019big}  & 0.162 & 11.466    &0.757 &0.913 &0.962

\\  \hline 
DepthFormer \cite{li2023depthformer}  &  0.152 &11.051  &0.689 &0.845 &0.947
\\  \hline 
PixelFormer\cite{agarwal2023attention}   & 0.151 &10.920  &0.691 &0.856 &0.949
\\ \hline 
NeWCRF\cite{yuan2022new}  & 0.219	&10.98 	&0.702 &0.881 &0.951
\\  \hline
BinsFormer\cite{li2024binsformer} & 0.149  &10.866 &0.732   & 0.894     &0.956
\\  \hline  
Adabins \cite{bhat2021adabins}  &  0.201  &10.240 & 0.748 &0.912 &0.962
\\  \hline 
IDisc\cite{piccinelli2023idisc}  &0.163  & 8.989  &0.809& 0.934 &0.971
\\  \hline 
Ours   &0.145  & 8.673       &0.823  &0.947   &0.980  
\\  \hline  
\vspace{-5mm}
\end{tabular}}
\end{center}
\vspace{-6.5mm}
\caption{ For a quantitative depth comparison  of the DDAD \cite{guizilini20203d}. }
\vspace{-3mm}
\label{DDAD_result}

\end{table}

\vspace{-1mm}


\vspace{-1mm}
\subsection{Monocular Depth Estimation Evaluation}
\vspace{-1mm}
For outdoor scenes, we evaluated our model using the standard KITTI Eigen split, which includes 697 images, and the DDAD \cite{guizilini20203d} dataset. Tables \ref{Kitti_result} and \ref{DDAD_result} summarize the performance of state-of-the-art supervised models on  KITTI and DDAD datasets, respectively, clearly showing that our method significantly outperforms existing approaches. Even compared to model architectures that also utilize textual descriptions, embodying depth information led to substantial improvements. specifically, on the KITTI dataset, our RMSE decreased to 1.654, and on the DDAD dataset, RMSE dropped to 8.673 , with all metrics achieving optimal results. As shown in Fig \ref{qualitative}, our model, supported by priors on object size and approximate shape, achieves better scale estimation than depth predictions by ECoDepth \cite{Patni_2024_CVPR}, MIM \cite{xie2023revealing}, and AFNet \cite{cheng2024adaptive} . By integrating scene depth and RGB image features, the model gains not only geometric priors but also visual details of the environment, enhancing depth estimation accuracy. Here, environmental text descriptions serve as another dimension of embodiment, enabling the model to recognize the presence of specific objects in the image. This knowledge simplifies the task to “plact” the object within the 3D scene based on its shape and spatial position. We demonstrate that scale can be inferred from language, effectively constraining the solution space for depth prediction and improving accuracy.
\begin{table}[t]

\begin{center}
\resizebox{0.48\textwidth}{!}{
\begin{tabular}{lllllllll}
\multicolumn{1}{c}{\bf ID} & 
\multicolumn{1}{c}{\cellcolor{pink}ESD} & 
\multicolumn{1}{c}{\cellcolor{pink}EFF} & 
\multicolumn{1}{c}{\cellcolor{pink}TD} & 
\multicolumn{1}{c}{\cellcolor{pink}DD} & 
\multicolumn{1}{c}{\cellcolor{blue!30}AbsRel ↓} & 
\multicolumn{1}{c}{\cellcolor{blue!30}Sq Rel↓} & 
\multicolumn{1}{c}{\cellcolor{blue!30}RMSE↓} & 
\multicolumn{1}{c}{\cellcolor{blue!30}$\delta < 1.25$\ ↑} 
\\ \hline 
1  &  & & &  &0.050 &0.139 &2.039   &0.976
\\  \hline 
2    & \checkmark &  & & &0.0310  &0.0620 &1.784    &0.988
\\  \hline 
3   &\checkmark &  \checkmark & &  &0.0314   &0.0648   &1.833    &0.984
\\  \hline  
4   &\checkmark & \checkmark &\checkmark& &0.0258    & 0.0453    & 1.698    & 0.990 
\\  \hline 
5   &\checkmark & \checkmark  & \checkmark & \checkmark &0.0251    & 0.0428   &  1.654   &0.991
\\  \hline  
\end{tabular}}
\end{center}
\vspace{-6mm}
\caption{Ablation study of our methods on the KITTI: ESD: Embodied Scene Depth. EFF: Embodied Feature Fusion. TD: Text Description. DD: Depth Description.  }
\vspace{-6mm}
\label{Ablation_result_model}
\end{table}


\subsection{Ablation study}
To thoroughly assess the impact of the proposed components in our methods on performance, we conducted detailed ablation studies on KITTI, presented in Table \ref{Ablation_result_model}.
\textbf{Embodied Scene Depth (ESD) :} Comparing Row 1 and Row 2, Embodied Scene Depth effectively enhances the model's depth accuracy for ground areas by combining camera physical properties with scene geometry. This initial depth significantly boosts the potential of embodiment depth in monocular depth estimation, showing outstanding performance in improving accuracy and detail capture for ground depth estimation.
\textbf{ Embodied Feature Fusio (EFF) :} Our analysis demonstrates that feature fusion is essential for improving depth estimation accuracy. Comparing the results in Row 2 and Row 3 reveals that integrating Embodied Scene Depth with RGB features significantly enhances the model’s depth estimation performance. Embodied Scene Depth provides geometric priors that strengthen spatial layout, while RGB features add visual details such as texture and color.
\textbf{Text Description (TD) :}  Our analysis shows that text descriptions play a crucial role in depth estimation. Comparing the results in Row 3 and Row 4 reveals that introducing text descriptions of the image as depth priors significantly improves depth estimation accuracy. Text descriptions provide scale information and semantic guidance for the scene, especially aiding in inferring object size and position. This additional information enhances the model’s adaptability and generalization across different perspectives and environments, thereby improving accuracy of depth estimation.
\textbf{Depth Description (DD) :} Incorporating depth information into text descriptions provides the model with additional relative depth cues and precise depth values for objects, significantly improving accuracy. Comparing the results in Row 3 and Row 4, it is evident that this quantitative depth data helps the model better understand object positioning within the scene, making depth estimation more aligned with real-world depth structures and enabling more reliable predictions in complex scenes.
\vspace{-2mm}

\section{Conclusion} 
\vspace{-2mm}
\label{sec:conclusion}
This study approaches depth estimation from the perspective of embodiment, integrating physical environmental information into a deep learning model so that perception and understanding depend not only on external data inputs but are also closely connected to the physical environment. Through interaction with road environments, we compute embodied scene depth and fuse it with RGB image features, providing the model with a comprehensive perspective that combines geometric and visual details. Also, we treat text descriptions that contain environmental context and depth information as another dimension of embodiment, introducing a object prior for scene understanding. Classical descriptions often struggle to accurately reflect spatial relationships between objects, but our approach incorporates depth clues, giving text descriptions a clear sense of depth hierarchy. This integration achieves accurate depth estimation within an embodiment framework.

\vspace{2mm}
\textbf{Acknowledgment}: 
This work is supported by NSF Awards NO. 2340882, 2334624, 2334246, and 2334690.
\vspace{-6mm}
{
    \small
    \bibliographystyle{ieeenat_fullname}
    \bibliography{main}
}


\end{document}